\documentclass[10pt,twocolumn,letterpaper]{article}

\usepackage{cvpr}              

\usepackage{graphicx}
\usepackage{amsmath}
\usepackage{amssymb}
\usepackage{booktabs}

\usepackage{enumitem}
\usepackage{multirow}
\usepackage{makecell}
\usepackage{bbding}
\usepackage{bm}

\usepackage[pagebackref,breaklinks,colorlinks]{hyperref}

\usepackage[capitalize]{cleveref}
\crefname{section}{Sec.}{Secs.}
\Crefname{section}{Section}{Sections}
\Crefname{table}{Table}{Tables}
\crefname{table}{Tab.}{Tabs.}

\def\cvprPaperID{*****} 
\def\confName{CVPR}
\def\confYear{2024}

\begin{document}

\title{Thin-Plate Spline-based Interpolation for Animation Line Inbetweening}

\author{Tianyi Zhu\textsuperscript{}, Wei Shang\textsuperscript{}, Dongwei Ren\textsuperscript{*}, Wangmeng Zuo\textsuperscript{}\\
\textsuperscript{}Harbin Institute of Technology, China\\
}
\maketitle

\begin{abstract}
Animation line inbetweening is a crucial step in animation production aimed at enhancing animation fluidity by predicting intermediate line arts between two key frames. 
However, existing methods face challenges in effectively addressing sparse pixels and significant motion in line art key frames. 
In literature, Chamfer Distance (CD) is commonly adopted for evaluating inbetweening performance. 
Despite achieving favorable CD values, existing methods often generate interpolated frames with line disconnections, especially for scenarios involving large motion.  
Motivated by this observation, we propose a simple yet effective interpolation method for animation line inbetweening that adopts thin-plate spline-based transformation to estimate coarse motion more accurately by modeling the keypoint correspondence between two key frames, particularly for large motion scenarios.
Building upon the coarse estimation, a motion refine module is employed to further enhance motion details before final frame interpolation using a simple UNet model.
Furthermore, to more accurately assess the performance of animation line inbetweening, we refine the CD metric and introduce a novel metric termed Weighted Chamfer Distance, which demonstrates a higher consistency with visual perception quality. Additionally, we incorporate Earth Mover's Distance and conduct user study to provide a more comprehensive evaluation.
Our method outperforms existing approaches by delivering high-quality interpolation results with enhanced fluidity. 
The code is available at \url{https://github.com/Tian-one/tps-inbetween}. 
\end{abstract}

\section{Introduction}
\label{sec:intro}

\begin{figure}[tb]
  \centering
  \setlength{\abovecaptionskip}{0pt} 
  \setlength{\belowcaptionskip}{0pt}
  \includegraphics[width=\linewidth]{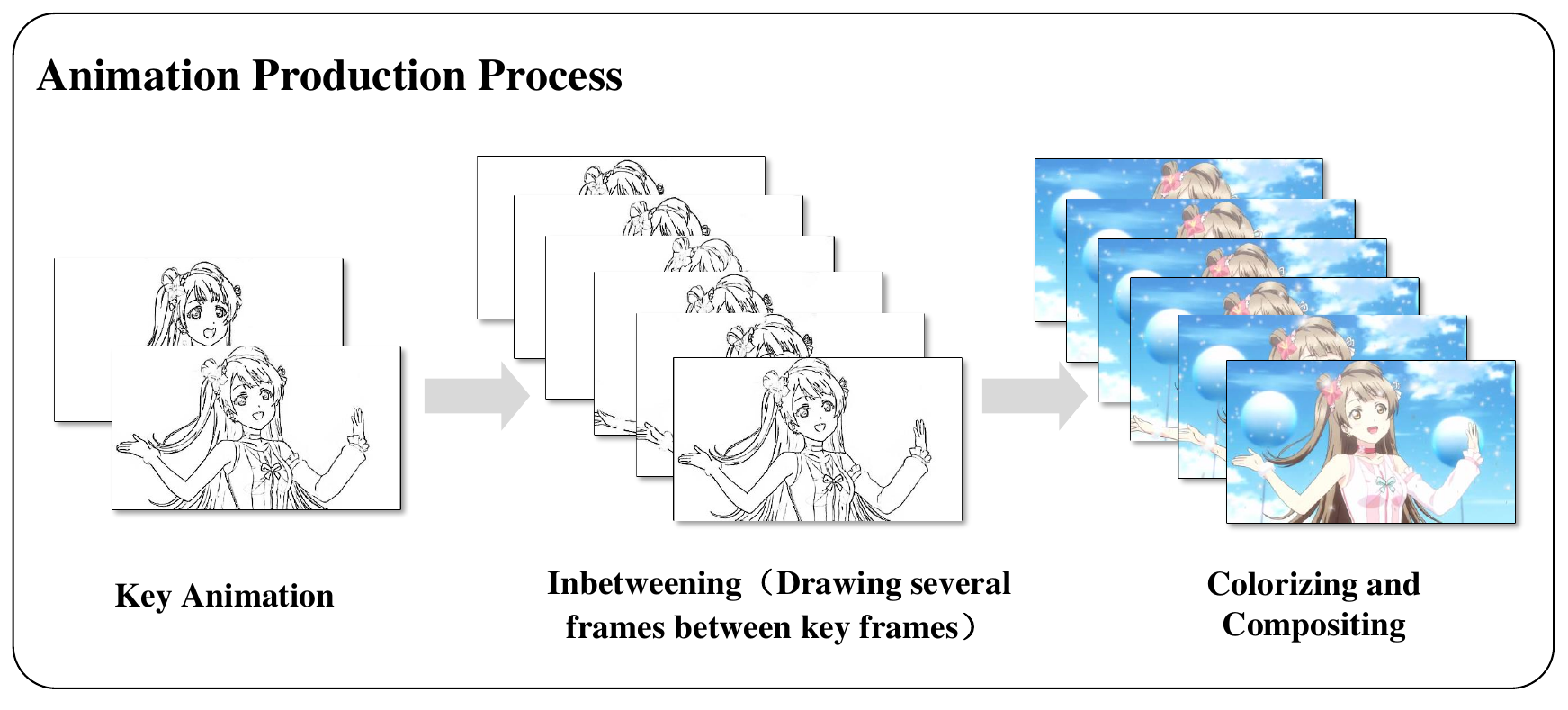}
  \caption{Common workflow of animation production, including creating key animation, inbetweening, colorizing and compositing.}
  \label{fig:anime_produce}
\end{figure}

2D animation plays a pivotal role in various domains, ranging from entertainment to education. 
With the advancement of generative models based on large language models, it may be feasible that 2D animation can be directly generated through video generation techniques \cite{guo2023animatediff,xing2024tooncrafter}. 
Despite involving conditions \cite{zhang2024controlvideo,mou2024t2i,zhang2023adding}, generating animation videos using given key frames provided by animators remains non-trivial. 
Therefore, it is still crucial to assist animators in completing animation videos by providing several key frames.
As shown in Fig. \ref{fig:anime_produce}, the animation creation process involves several distinct steps: (i) Creating key animation frames by animators,
(ii) Inbetweening, responsible for interpolating intermediate frames, 
and (iii) Colorizing and compositing the line arts. 
In previous studies, inbetweening and colorization have been sometimes used interchangeably.
For example, in \cite{siyao2021deep, chen2022improving}, inbetweening is addressed through video interpolation after colorizing and compositing key frames. 
Although automatic colorization techniques \cite{zhang2021user, li2022eliminating} offer certain advantages, they compromise the visual quality during the interpolation process, and present challenges for animators in terms of reprocessing post-colorized frames.

Therefore, this work devotes attention to animation line inbetweening, \textit{i.e.}, frame interpolation on line arts of key frames. 
This task presents challenges due to the presence of substantial motion amplitudes and sparse pixel distribution, rendering existing video interpolation methods unsuitable \cite{kalluri2023flavr, huang2022real}. To overcome the large motion of line arts, in \cite{siyao2023deep}, a graph-based approach is employed to vectorize and reposition vertices of the line art for generating intermediate graphs. 
While graph-based representation facilitates cleaner lines, it also introduces artifacts and often results in discontinuous lines that affect interpolation outcomes. For raster images, inaccuracies in vectorization methods can also impact the performance of this approach.

In this work, we propose a novel raster image-based interpolation pipeline for animation line inbetweening, as shown in Fig. \ref{fig:model}. We first introduce the thin-plate spline (TPS) transformation \cite{bookstein1989principal} to model coarse motion, by which the intermediate poses particularly for large motion can be more accurately obtained by using keypoints correspondence priors.
Building upon the coarse estimation, a simple motion refine module is proposed, consisting of an optical flow network for refining coarse motion and a simple UNet model for synthesizing final frames.
Additionally, we introduce an improved metric dubbed Weighted Chamfer Distance (WCD), which  exhibits a closer alignment with human visual perception compared to the original CD metric \cite{chen2022improving,siyao2023deep}. 

Extensive experiments are conducted on benchmark datasets, and our method is compared against state-of-the-art techniques in video interpolation \cite{huang2022real, zhang2023extracting, li2023amt}, animation interpolation \cite{chen2022improving}, and line inbetweening \cite{siyao2023deep}. 
The evaluation metrics include CD and WCD scores, as well as introducing Earth Mover’s Distance (EMD) and user study into consideration. 
Our approach outperforms existing methods by producing high-quality interpolation results with enhanced fluidity for all three interpolation gaps, \textit{i.e.}, 1, 5, and 9.
Our contributions can be summarized as:
\begin{itemize}[itemsep=2pt,topsep=2pt,parsep=0pt]
    \item We propose a simple but effective framework for animation line inbetweening, which incorporates a TPS-based module to estimate coarse motion, and a motion refine module for refining motion details and generating final frames.
    \item The thin-plate spline-based interpolation module is proposed for better modeling the keypoint correspondence to handle large motion and line disconnections.  
    \item We present an improved metric WCD and introduce the EMD metric for evaluating inbetweening effect.  
    Extensive experiments demonstrate that our approach outperforms existing methods by generating high-quality interpolation results with enhanced fluidity.
\end{itemize}

\begin{figure*}[tb]
  \centering
    \setlength{\abovecaptionskip}{0pt} 
  \setlength{\belowcaptionskip}{0pt}
  \includegraphics[width=\linewidth]{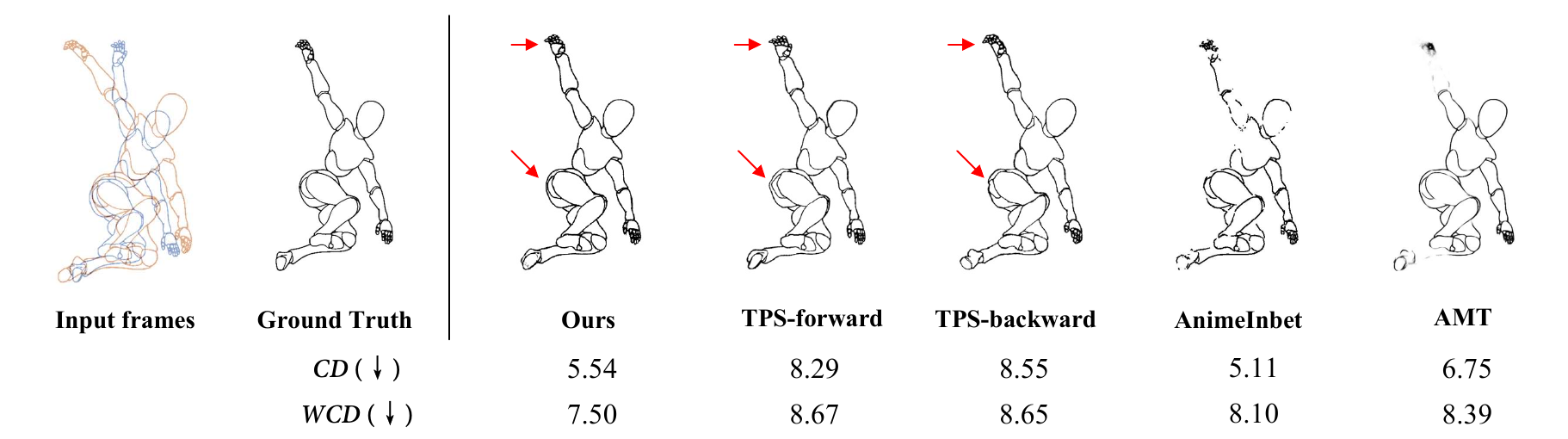}
  \caption{Visual comparison and corresponding metrics. The left side showcases the overlapped input frames along with the ground truth, while the right side displays the interpolation results and their evaluation metrics of inbetweening methods. TPS-forward and TPS-backward refer to the results obtained only using the flow estimated by the TPS module to warp the input's first and last frames, respectively. WCD is an enhanced metric that we proposed, which is better aligned with human visual perception, as detailed in Eq.~\eqref{eq:WCD}.
  }
  \label{fig:metric vis}
\end{figure*}

\section{Related Work}
\noindent {\bf Video Frame Interpolation.} In recent years, numerous methods have been proposed to address the challenge of video frame interpolation \cite{jiang2018super, shang2023joint, sim2021xvfi, reda2022film, kalluri2023flavr, yoo2023video, wu2023boost, Hu_2024_CVPR}. Some methods attempted to extract features from concatenated input frames to learn kernels \cite{niklaus2017video, lee2020adacof} or directly synthesize intermediate frames \cite{kalluri2023flavr}. The most recent state-of-the-art techniques \cite{li2023amt, zhang2023extracting, wu2024perception} predominantly leverage optical flow. They predict the optical flow between the intermediate frames and the two input frames, utilizing warping to obtain the intermediate frames. Liu \textit{et al.} \cite{Liu_2024_CVPR} introduce sparse global matching to refine optical flow for large motion. RIFE \cite{huang2022real} adopts a privileged distillation scheme to directly approximate the intermediate flows for real-time frame interpolation. Zhang \textit{et al.} \cite{zhang2023extracting} proposed a hybrid CNN and Transformer \cite{vaswani2017attention} architecture to extract motion and appearance information. Li \textit{et al.} \cite{li2023amt} introduced bidirectional correlation volumes to obtain fine-grained flow fields from updated coarse bilateral flows. While these methods have demonstrated superior performance in video frame interpolation within real-world scenarios, their application to the line art inbetweening often encounters challenges. The sparse pixel distribution in line art images and substantial inter-frame motion frequently result in undesired effects, such as blurring and artifacts.

\noindent {\bf Line Art Inbetweening.} Previous works on line art inbetweening were often conducted under strict conditions \cite{yang2017context} or employed methods similar to video frame interpolation \cite{shen2023bridging, narita2019optical}. However, these approaches consistently faced challenges in handling large motion. Recently, Siyao \textit{et al.} \cite{siyao2023deep} attempted to address line art inbetweening by converting line art into the graph, then repositioning vertex to synthesize an intermediate graph. Although this method considers the case of large motion, several issues persist: 1) For raster images, the accuracy of vectorization significantly impacts the interpolation results. 2) The correspondence of the vectorized graph vertices is crucial when training, which is challenging to obtain on wild animation. 3) Discontinuous lines sometimes appear due to the mask of vertices. Compared with them, our model is more flexible based on raster images, and the two-stage pipeline design can allow manual fine-tuning to achieve higher quality results.

\noindent {\bf Video Generation.} There are many researches based on diffusion models \cite{rombach2022high} obtained remarkable results on video generation. Some approaches leverage text or a combination of text and images \cite{singer2023makeavideo, zeng2023make, guo2023animatediff} to generate video. While some research is dedicated to exploring how to generate videos with control \cite{zhang2024controlvideo,mou2024t2i,zhang2023adding}, producing stable and smooth video sequences remains challenging, particularly when dealing with the large motion. Moreover, the transference of conventional video generation techniques to the domain of line art inbetweening is further complicated due to the domain gap. For animation, plot arrangement, the density of the rhythm, and the quality of inbetween frames are paramount. Consequently, our focus is on assisting animators with creating inbetween drawings rather than directly generating the final animation.

\begin{figure}[!t]
  \centering
    \setlength{\abovecaptionskip}{0pt} 
  \setlength{\belowcaptionskip}{0pt}
  \includegraphics[width=\linewidth]{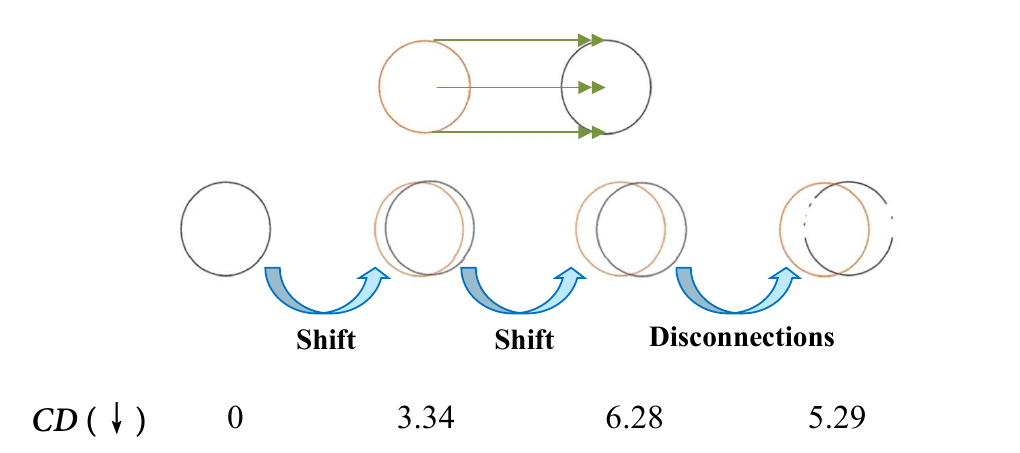}
  \caption{A simple evaluation case of orange and black circular objects. 
  In the presence of line disconnections, the CD metric is misleading.
  }
  \label{fig:metric}
\end{figure}

\section{Proposed Method}

In this section, we first discuss the limitations of the existing evaluation metrics and methods \cite{siyao2023deep, shen2023bridging}, and then propose a simple yet effective pipeline to solve large motion of inbetweening for raster line art images.

\subsection{Limitations in Existing Methods and Metric}
\label{sec:limit}
To obtain better inbetween quality, a top challenge is overcoming exaggerated and non-linear large motions of line art. Previous works on real-world video frame interpolation \cite{sim2021xvfi, reda2022film} attempted to directly use optical flow and multi-scale strategy to handle large motion. However, these approaches always failed on line art inbetweening because of the sparse pixels of the line art, which make it difficult to obtain accurate optical flow and context information. 

 Previous works on line art inbetweening \cite{siyao2023deep, narita2019optical} have mainly adopted Chamfer Distance (CD) \cite{chen2022improving} to evaluate the quality of the results, computed as 

\begin{equation}
  {CD}\left(\bm{b}_0, \bm{b}_1\right) = \frac{1}{2HWD}\sum (\bm{b}_0 \odot\mathcal{D}(\bm{b}_1) + \bm{b}_1\odot\mathcal{D}(\bm{b}_0)),
  \label{eq:CD}
\end{equation}
where $\bm{b}_0$ and $\bm{b}_1$ are binary images with effective pixels is $1$, $\mathcal{D}$ is the distance transform, and $\bm{b}_0 \odot\mathcal{D}(\bm{b}_1)$ denotes the distance map, wherein each effective pixel in $\bm{b}_0$ is labeled with the distance to the nearest effective pixel in $\bm{b}_1$, $\odot$ is element-wise multiplication, $H$, $W$, $D=\sqrt{H^2 + W^2}$ are height, width, and diameter, respectively. 

Although these methods achieve favorable CD scores, we still observe significant issues with line disconnections as shown in \cref{fig:metric vis}. This raises the question: can CD truly reflect the quality of line art inbetweening? Let's consider a straightforward motion scenario: a circle is translated from left to right, as shown in \cref{fig:metric}. When the predicted circle aligns with the ground truth, the value of CD is $0$. As the predicted results shift gradually, the CD value increases, which is expected. However, if random lines are erased from the circle, the CD value decreases, which means the quality is better. This is obviously unreasonable. The main reason for the error in quality evaluation is that CD simply averages the sum of $\bm{b}_0 \odot\mathcal{D}(\bm{b}_1)$ and $\bm{b}_1\odot\mathcal{D}(\bm{b}_0)$, without considering the count of effective pixels. 

To overcome the limitations of the existing metric, we introduce the perceptually aligned WCD metric and augment our evaluation with the EMD metric. Details are provided in the evaluation metrics section.
To address the limitations of existing methods, 
we propose an effective two-stage approach for 2D hand-drawn animation inbetweening, mimicking animators' practice of anchoring motion with key features. Initially, we employ keypoint matching to detect local feature correspondences, followed by TPS for motion modeling, offering an initial estimate sensitive to large motion as shown in \cref{fig:metric vis}. TPS's plane deformation preserves line continuity. This module yields aligned frames that are subsequently refined by a simple network for precise motion tuning.

\begin{figure*}[tb]
  \centering
    \setlength{\abovecaptionskip}{0pt} 
  \setlength{\belowcaptionskip}{0pt}
  \includegraphics[width=0.9\textwidth]{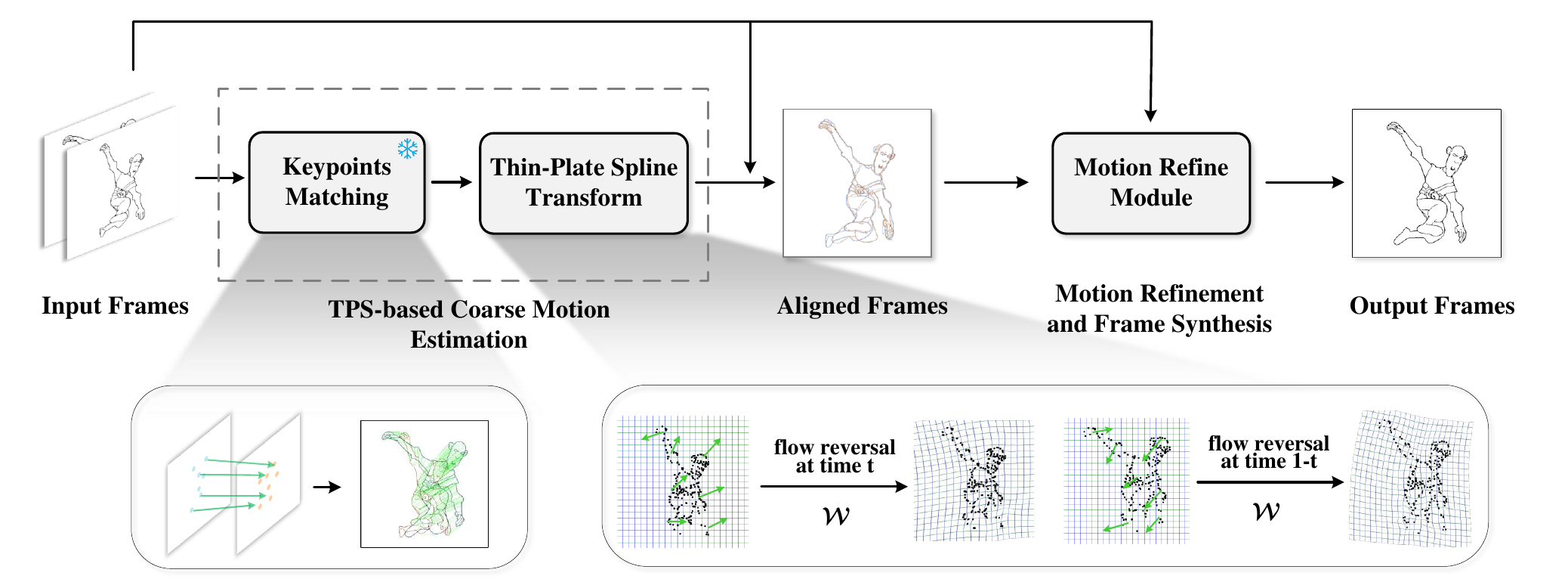}
  \caption{The overall pipeline of our approach begins with the initial estimation of coarse motion using a TPS-based transformation, which models point correspondence between key frames. Subsequently, we employ a simple motion refine module consists an optical flow network for estimating fine motion and a simple UNet model for synthesizing the inbetweening results.}
  \label{fig:model}
\end{figure*}

\subsection{Thin-Plate Spline-based Inbetweening Method}

Given two input raster line art images $\bm{y}_0$ and $\bm{y}_1$, where $\bm{y}_0, 
 \bm{y}_1\in\mathbb{R}^{H\times W}$, and $H$ and $W$ are the height and the width, respectively.
The goal of our method is to interpolate an intermediate raster line art image $\bm{y}_t$ with $\bm{y}_t \in\mathbb{R}^{H\times W}$ at arbitrary time $t \in (0,1)$. 
The overall pipeline of our method is illustrated in Fig.~\ref{fig:model}.

\subsubsection{TPS-based Coarse Motion Estimation.} 
Due to non-linear large motion in line art inbetweening, we employ a coarse motion estimation approach based on TPS to predict the motion instead of directly utilizing optical flow. 
First, we feed $\bm{y}_0$ and $\bm{y}_1$ into a pre-trained keypoints matching model to extract keypoints and establish the correspondence of each keypoint $\{(x^i_0, y^i_0), (x^i_1, y^i_1)\}_{i=1}^{n}$,
where $(x^i_0, y^i_0)$ and $(x^i_1, y^i_1)$ represent coordinates of the $i$-th valid keypoint pair in $\bm{y}_0$ and $\bm{y}_1$, respectively.
In the following, we use $p^i_0$ and $p^i_1$ to represent $(x^i_0, y^i_0)$ and $(x^i_1, y^i_1)$ for simplicity.
We in this paper apply pre-trained keypoints matching model and freeze it during training.
To utilize the correspondence of keypoint pairs $\{p^i_0, p^i_1\}_{i=1}^{n}$ for predicting the rough motion between line art images, we draw inspiration from the thin-plate spline (TPS) transformation \cite{bookstein1989principal}.
TPS transformation addresses the problem of interpolating surfaces over scattered data and modeling shape change as deformation. It utilizes thin-plate splines to interpolate surfaces based on a fixed set of points in the plane.
Specifically, given a set of corresponding keypoint pairs$\{p^i_0, p^i_1\}_{i=1}^{n}$, we can directly use the TPS transformation function \cite{bookstein1989principal} to model the motion between the two sets of keypoints, represented as
\begin{equation}
    \mathcal{F}_{0\rightarrow 1}(p_1)=\bm A \begin{bmatrix} {p}^\mathrm{T}_1 \\ 1 \end{bmatrix} + \sum^{n}_{i=1}w_{i}\mathcal{U}\left(\left\| p^{i}_0-p_1 \right\|_2\right)
    \label{eq:tps function}
\end{equation}  
where $p_1 \in \{p^i_1\}^{n}_{i=1}$ is an arbitrary keypoint coordinate of $\bm{y}_1$, $\bm A\in \mathbb{R}^{2\times3}$ means affine transformation matrix on linear space, $n$ is the number of keypoints, $w_{i}\in \mathbb{R}^{2\times1}$ is the weight of the $i$-th keypoint of $p^i_0$, and $\mathcal{U}(k)=k^{2}\log (k)$ is the thin-plate spline kernel, which represents a mapping to displacement between the control point $p^i_0$ and the final point $p_1$.
The energy function $\mathcal{E}$ of TPS formula is minimized by
\begin{equation}
  \mathcal{E} = {\iint_{\mathbb{R}^2}\left(\left(\frac{\partial^{2}\mathcal{F}}{\partial^{2}x^2}\right)^{2}+2\left(\frac{\partial^{2}\mathcal{F}}{\partial x\partial y}\right)^{2} + \left(\frac{\partial^{2}\mathcal{F}}{\partial^{2}y^2}\right)^{2}\right)dxdy}.
  \label{eq:tps bending energy}
\end{equation}
To solve this optimization problem, we employ a linear system to compute the values of $\bm A$ and $\bm W$:
\begin{equation}
  \begin{bmatrix} \bm K & \bm P \\ \bm P^\mathrm{T} & \bm O \end{bmatrix} \begin{bmatrix} \bm W^\mathrm{T} \\ \bm A^\mathrm{T} \end{bmatrix} = \begin{bmatrix} \bm v \\ \bm o \end{bmatrix},
  \label{eq:tps solver}
\end{equation}
where $\bm K_{ij}=\mathcal{U}\left(\left\|p^{i}_{0}-p^{j}_{0}\right\|_2\right)$, $\bm K \in\mathbb{R}^{n\times n}$, $i$-th row of $\bm P$ is $[p^{i}_{0},1]$, $\bm v\in\mathbb{R}^{n\times 2}$ is a column vector composed of $\{p^i_1\}^{n}_{i=1}$, $\bm O\in\mathbb{R}^{3\times 3}$ and $\bm o\in\mathbb{R}^{3\times 2}$ are zero matrices. 

\begin{figure*}[tb]
  \centering
    \setlength{\abovecaptionskip}{0pt} 
  \setlength{\belowcaptionskip}{0pt}
  \includegraphics[width=0.9\textwidth]{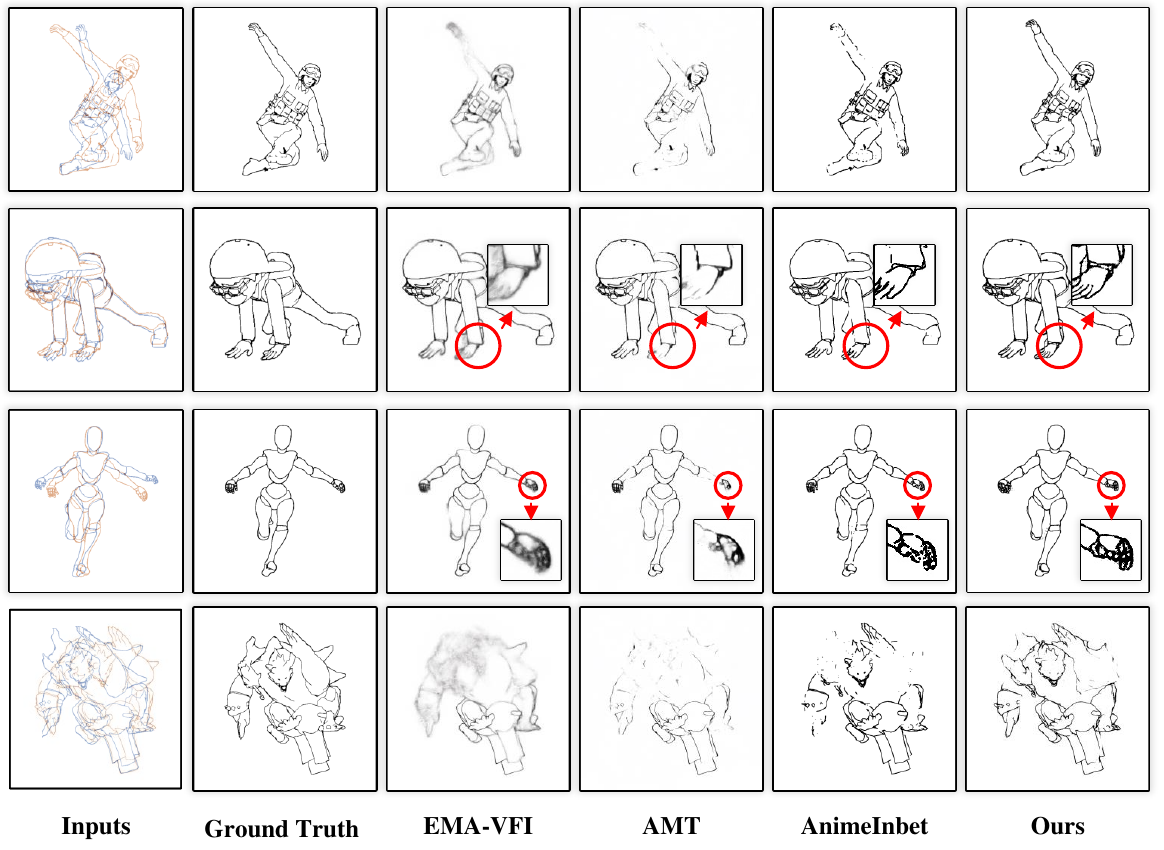}
  \caption{Inbetweening results comparison on the MixiamoLine240 dataset. }
  \label{fig:compare}
\end{figure*}

After obtaining the affine matrix $\bm A$ and weight matrix $\bm W$ in TPS, we initialize the coordinate grid $\bm g\in \mathbb{R}^{H\times W \times 2}$ and apply it to Eq.~\eqref{eq:tps function} to obtain the coarse motion map $\bm m_{0\rightarrow1}$ by
\begin{equation}
  \bm m_{0\rightarrow1} = \mathcal{F}_{0\rightarrow 1}(\bm g).
  \label{eq:motion from tps}
\end{equation}
Similarly, we can also obtain the coarse motion flow $\bm m_{1\rightarrow0}$ from $\bm{y}_1$ to $\bm{y}_0$.

Now we get the coarse motion between two frames, and our goal is to interpolate intermediate frame $\bm{y}_t$ at time $t \in (0,1)$. 
Since $\bm{y}_t$ is unavailable, we cannot directly compute the motion $\bm m_{t\rightarrow0}$ and $\bm m_{t\rightarrow1}$.
Drawing inspiration from \cite{jiang2018super}, we utilize linear models based on the assumption that the optical flow field is locally smooth to approximately compute the 
intermediate motion flow by
\begin{equation}
    \begin{aligned}
    \bm m_{t\rightarrow0} = t \bm m_{1\rightarrow0} &= t ((1-t)\bm m_{1\rightarrow0} + t \bm m_{1\rightarrow0}) \\
    &= -(1-t)t \bm m_{0\rightarrow1} + t^2 \bm m_{1\rightarrow0}.
    \end{aligned}
  \label{eq:flow reversal}
\end{equation}
Similarly, we can obtain the motion $\bm m_{t\rightarrow1}$.

The coarsely aligned frames are obtained through backward warping operation $\mathcal{W}$:
\begin{equation}
  \tilde{\bm{y}}_{0\rightarrow t} = \mathcal{W}\left(\bm{y}_0, \bm m_{t\rightarrow0}\right), \tilde{\bm{y}}_{1\rightarrow t} = \mathcal{W}\left(\bm{y}_1, \bm m_{t\rightarrow1}\right).
  \label{eq:warp}
\end{equation}

\subsubsection{Motion Refinement Module} 
After performing the coarse alignment operation, we have reduced the motion between the two frames.
At this stage, we feed the coarsely aligned frames into optical flow methods to further refine the motion.
We employ a lightweight and efficient module IFBlock \cite{huang2022real} to iteratively compute the residual motion flows $\Delta\bm m_{t\rightarrow1}$, $\Delta\bm m_{t\rightarrow0}$ and fusion map $\bm f$:
\begin{equation}
\begin{split}
  &\Delta\bm m^{\alpha}_{t\rightarrow1}, \Delta\bm m^{\alpha}_{t\rightarrow0}, \bm f^{\alpha} =  \\ 
  &\mathrm{IFBlock}_{\alpha}\left(\tilde{\bm{y}}_{0\rightarrow t}, \tilde{\bm{y}}_{1\rightarrow t}, \Delta\bm m^{\alpha-1}_{t\rightarrow1}, \Delta\bm m^{\alpha-1}_{t\rightarrow0}, \bm f^{\alpha-1}\right),
  \label{eq:IFNet} 
\end{split}
\end{equation}
where $\mathrm{IFBlock}_{\alpha}$ is the $\alpha$-th IFBlock. 
When $\alpha$ is equal to $1$, we directly pass $\tilde{\bm{y}}_{0\rightarrow t}$ and $\tilde{\bm{y}}_{1\rightarrow t}$ into the IFblock.
The total number of IFBlocks is $3$ in our method, and we use the output in the last iteration as the residual motion flows and fusion map.
Then we add the residual motion flows to coarse motion flow yielding refined motion flows $\bm s_{{t\rightarrow0}}$ and $\bm s_{{t\rightarrow1}}$.
We can obtain the refinedly aligned frames $\hat{\bm{y}}_{0\rightarrow t}$ and $\hat{\bm{y}}_{1\rightarrow t}$ with refined motion flows.
\begin{equation}
  \hat{\bm{y}}_{0\rightarrow t} = \mathcal{W}\left(\bm{y}_0, \bm s_{t\rightarrow0}\right),
  \hat{\bm{y}}_{1\rightarrow t} = \mathcal{W}\left(\bm{y}_1, \bm s_{t\rightarrow1}\right).
\end{equation}
By combining the two refinedly aligned frames, weighted by time $t$, we can obtain the initial inbetweening result $\hat{\bm{y}}_{t}$.
Moreover, we utilize a multi-scale feature alignment approach where, for each scale, we warp the features extracted from $\bm{y}_0$ and $\bm{y}_1$ to obtain multi-scale aligned features $\{\bm{y}^{l}_0\}^{L}_{l=1}$ and $\{\bm{y}^{l}_1\}^{L}_{l=1}$. And we set $L=4$ in our experiments.

Finally, we utilize a simple UNet \cite{ronneberger2015u} for predicting the residual inbetweening, whose input is the concatenation of $\bm{y}_0, \bm{y}_1, \hat{\bm{y}}_{0\rightarrow t}, \hat{\bm{y}}_{1\rightarrow t}, \bm s_{{t\rightarrow0}}, \bm s_{{t\rightarrow1}}, \bm f,\{\bm{y}^{l}_0\}^{L}_{l=1}, \{\bm{y}^{l}_1\}^{L}_{l=1}$.
The UNet architecture consists of $(L+1)$ scales, from $0$ to $L$. At the $0$-th scale, the concatenation of input is first processed for feature extraction. 
For subsequent scales, features in $l$-th scale of UNet are concatenated with the $l$-th warped features $\bm{y}^{l}_0$ and $\bm{y}^{l}_1$ and then fed into the $l$-th encoder layer.
Finally, we add the residual inbetweening result $\Delta \hat{\bm{y}}_{t}$ to the initial inbetweening result $\hat{\bm{y}}_{t}$ to obtain final interpolation result ${\bm{y}}_{t}$.

\begin{table*}[tb]
 \setlength{\abovecaptionskip}{0pt} 
  \setlength{\belowcaptionskip}{0pt}
  \caption{Quantitative comparison (CD$\downarrow$ / WCD$\downarrow$ / EMD$\downarrow$) with state-of-the-art methods on different frame gaps. The first place and runner-up are highlighted in bold and underlined, respectively. 
  CD is scaled by $10^{5}$, WCD by $10^{4}$, EMD by $10^3$.
  }
  \label{tab:compare}
  \centering
   \small
  \begin{tabular}{@{}p{1.5cm}>{\centering\arraybackslash}p{2.0cm}>{\centering\arraybackslash}p{2.0cm}>{\centering\arraybackslash}p{2.3cm}>{\centering\arraybackslash}p{2.3cm}>{\centering\arraybackslash}p{2.3cm}>{\centering\arraybackslash}p{2.3cm}@{}}
    \toprule
    Method & \multicolumn{2}{c}{$N=1$ (validation / test)} & \multicolumn{2}{c}{$N=5$ (validation / test)} & \multicolumn{2}{c}{$N=9$ (validation / test)}\\
     \specialrule{.05em}{.4ex}{.65ex}
     EMA-VFI & 3.01/8.62/5.84 & 3.70/8.59/6.16 & 15.21/12.04/11.1 & 17.47/12.00/11.31 & 28.44/14.94/13.95 & 31.30/14.86/14.05\\
    RIFE & 3.43/8.28/5.48 & 2.93/8.39/3.81 & 14.14/11.27/8.37 & 15.82/11.15/9.63 & 25.70/13.67/11.61 & 28.27/13.59/11.77\\
    AMT & \underline{2.43}/\underline{7.33}/\underline{2.46} & \underline{2.84}/\underline{7.31}/\underline{2.34} & 14.34/9.58/3.83 & 16.65/10.13/4.07 & 22.39/11.62/5.02 & 25.77/12.21/5.24\\
     PerVFI & 2.88/7.91/2.98 & 3.32/7.79/2.78 & 11.90/9.23/\underline{3.29} & 12.62/\underline{9.05}/\underline{3.37}& 19.80/10.20/\underline{3.42} & 21.38/\underline{10.09}/\underline{3.65}\\
    \specialrule{.05em}{.4ex}{.65ex}
     EISAI & 3.58/8.91/6.26 & 4.02/8.56/5.86 & 13.06/9.80/6.29 & 14.46/9.57/5.80 & 22.41/10.69/6.27 & 24.46/10.54/5.79\\

    \specialrule{.05em}{.4ex}{.65ex}
    AnimeInbet & 2.51/7.54/3.02 & 3.07/7.93/3.74 & \underline{9.33}/\underline{8.53}/3.51 & \textbf{10.74}/9.84/4.91 & \underline{16.08}/\underline{10.01}/4.36 & \textbf{17.76}/11.54/5.74\\
    Ours & \textbf{2.34}/\textbf{7.32}/\textbf{1.01} & \textbf{2.77}/\textbf{7.28}/\textbf{2.24} & \textbf{9.24}/\textbf{7.99}/\textbf{1.34} & \underline{10.79}/\textbf{8.39}/\textbf{2.88} & \textbf{15.89}/\textbf{8.85}/\textbf{1.69} & \underline{18.09}/\textbf{9.48}/\textbf{3.37}\\
  \bottomrule
  \end{tabular}
\end{table*}

\subsection{Loss Functions}
Most of the video frame interpolation works employ $\mathcal{L}_1$ loss to optimize PSNR \cite{chen2022improving, zhang2023extracting}. However, for line art, pixel-wise measurement is not suitable as slight misalignment can result in significant pixel-wise differences. Thus we use distance transform $\mathcal{D}$ to convert a line art frame into a distance transform map. The implementation involves using a differentiable convolutional distance transform layer \cite{pham2021differentiable}.
By minimizing the difference between the distance transform map of predicted frames and the ground truth, we ensure the similarity of the line structure. The loss is represented as
\begin{equation}
    \mathcal{L}_{dt} = \frac{1}{N} \sum_{t} \left\|\mathcal{D}(\bm{y}_t)-\mathcal{D}(\bm{y}_t^{gt})\right\|,
    \label{eq:dt loss}
\end{equation}
where $N$ is the interpolated frame number between two key frames.   

We noticed that better inbetweening results usually yield a sum of effective pixels roughly equivalent to that of the ground truth. Thus, we introduce a loss $\mathcal{L}_{cnt}$ to constrain the sum of effective pixels
\begin{equation}
   \mathcal{L}_{cnt} = \frac{1}{N} \sum_{t}  \left\|\sum\mathrm{ReLU}(\bm{y}_t - \eta) - \sum\mathrm{ReLU}(\bm{y}_t^{gt} - \eta )\right\|,
    \label{eq:cnt loss}
\end{equation}
where $\eta$ is the threshold to delimit the effective pixels.

To make the resulting line art clearer, we add binarization loss \cite{xie2021seamless} $\mathcal{L}_{bi}$ to encourage the network to generate black or white pixels
\begin{equation}
   \mathcal{L}_{bi} = \frac{1}{N} \sum_{t}  \left\| \left|\bm{y}_t - 0.5\right| - 0.5\right\|_2.
    \label{eq:bi loss}
\end{equation}
Further, we adopt LPIPS \cite{zhang2018unreasonable} loss $\mathcal{L}_{lpips}$ to improve the perceptual quality of line art.
The total loss $\mathcal{L}$ is defined as
\begin{equation}
   \mathcal{L} = \mathcal{L}_{dt}  + \lambda_{cnt} \mathcal{L}_{cnt} + \lambda_{bi} \mathcal{L}_{bi} + \lambda_{lpips} \mathcal{L}_{lpips},
    \label{eq:all loss}
\end{equation}
where $\lambda_{cnt}$, $\lambda_{bi}$, $\lambda_{lpips}$ are hyperparameters.

\section{Experiments}
\label{sec:Experiments}

\subsection{Implementation Details}
We implemented our model in PyTorch \cite{paszke2019pytorch}. We apply GlueStick \cite{pautrat2023gluestick} as our keypoints matching model. The training and testing were conducted on MixiamoLine240 dataset \cite{siyao2023deep}, a line art dataset with ground truth geometrization and vertex matching labels. It's worth noting that during training, we utilized only raster images from the dataset. We set the frame gap $N=5$ during training, and tested on the test set with the gaps $N=1,5,9$ respectively. For augmentation, we applied random temporal flipping. Our model was trained at the resolution of $512\times512$, and tested at the original resolution $720\times720$. We employed the Adam \cite{KingBa15} optimizer with $\beta_{1}=0.9$ and $\beta_2 = 0.999$ at a learning rate of $1\times10^{-4}$ for $50$ epochs. The training and testing were performed on an NVIDIA RTX A6000 GPU. The hyperparameters $\lambda_{lpips}$, $\lambda_{cnt}$, $\lambda_{bi}$ and $\eta$ were set as 5, 5, $1\times10^{-3}$ and $0.9$, respectively.

\subsection{Evaluation Metrics}
\label{sec:eval metrics}
PSNR and SSIM \cite{wang2004image} are common evaluation metrics in video interpolation tasks but they often fall short when applied to animation videos \cite{chen2022improving}.
Previous works have adopted CD as the evaluation metric. However, as discussed in the previous section, 
it has certain limitations. While it can reflect the alignment with the ground truth, it is affected by discontinuous lines.

To address the shortcomings of the inappropriate metric, we introduce an improved metric based on CD for evaluating line art visual quality, called WCD, aiming for better alignment with human perception.
Based on the observation that when the line art image maintains a similar overall structure to the ground truth and the lines remain continuous, the effective pixels should be approximately equal to those in the ground truth, we introduce a weighted factor $\mathcal{H}(\bm{b}_0 , \bm{b}_1) \in [0.5, 1)$  derived from the difference between the effective pixels of the two line art images.
Furthermore, we apply a non-linear softplus function mapping $\mathcal{G(\cdot)}$ to the values obtained from the distance transform, imposing a greater penalty on values with larger distance.
WCD can be represented as

\begin{equation}
\begin{split}
   {WCD}\left(\bm{b}_0, \bm{b}_1\right) = \frac{\mathcal{H}(\bm{b}_0 , \bm{b}_1 )}{HWD}&\sum (\mathcal{G}\left(\bm{b}_0 \odot\mathcal{D}(\bm{b}_1)\right) \\ &+ \mathcal{G}\left(\bm{b}_1\odot\mathcal{D}(\bm{b}_0)\right)),
    \label{eq:WCD}
\end{split}
\end{equation}

\begin{equation}
\begin{split}
  \mathcal{H}(\bm{b}_0 , \bm{b}_1 ) &= \mathrm{Sigmoid}\left(\frac{\left|\left| \bm{b}_0 \right|_n - \left| \bm{b}_1 \right|_n \right|}{\mathrm{min}\left(\left| \bm{b}_0 \right|_n, \left| \bm{b}_1 \right|_n\right)}\right)
\end{split}
\end{equation}
where $|\cdot|_n$ means the number of effective pixels.

As distance map only considers the nearest neighbour of a point, it falls short in adequately representing the distribution of these points. Therefore, we advocate for the adoption of EMD \cite{nguyen2021point, achlioptas2018learning}, a metric commonly used in 3D point cloud representation, to assist in the assessment of line art quality. 
Considering the potential computational cost of EMD calculation \cite{pele2009fast}, we compute EMD separately for the horizontal and vertical directions and then average the results.

\subsection{Comparison with State-of-the-arts}
We categorized the methods for comparison into three groups. For real-word video frame interpolation, we choose RIFE \cite{huang2022real}, EMA-VFI \cite{zhang2023extracting}, AMT \cite{li2023amt} and PerVFI \cite{wu2024perception}. For animation interpolation and line art inbetweening, we choose EISAI \cite{chen2022improving} and AnimeInbet \cite{siyao2023deep}, respectively. The AnimeInbet is a vector-based method, while the others are raster image-based methods. We finetune these methods with the frame gap $N=5$. We tested each model on validation set and test set with frame gaps of 1, 5, and 9. For a frame gap $N$,  evaluation metrics are based on the average of these generated $N$ frames. Predicted raster images were converted to binary by setting pixels smaller than $0.95$ to $0$. 

As shown in \cref{tab:compare}, with the frame gap of $1$,
the method of video frame interpolation based on optical flow \cite{li2023amt} can also achieves commendable performance. However, as the motion gradually increases, the performance of these interpolation models based on real-world videos significantly degrades. Though previous work focused on line art inbetweening \cite{siyao2023deep} has demonstrated some potential in handling large motions, our method outperforms on both test and validation sets.

In \cref{fig:metric vis}, a line art inbetweening result and corresponding evaluation metrics of our method are compared with other approaches. It can be observed that AnimeInbet achieved the highest score in CD, despite the presence of noticeable missing lines. Although our method not exhibit the highest performance in CD, it attains the best result in WCD that is more aligned with human visual perception. 

More inbetweening results are illustrated in \cref{fig:compare}. The two leftmost columns depict the overlapped inputs and the ground truth. To highlight details, certain parts of the image are enlarged and marked with the red circle.
In instances of simple motion, as exemplified in the second and third lines, our method outperforms other approaches in recovering details and preserving a more complete line structure. When the motion is large, EMA-VFI tends to blur, and AMT causes large areas of lines to disappear. Although AnimeInterp can restore the general outline, it often generates discontinuous lines. Compared with them, our approach produces more continuous and clear lines by leveraging modeled point correspondence. For complex line art inputs and motion with occlusion, as exemplified in the last line, our method also demonstrates robustness. More results and videos are available in the supplementary file. 

Additionally, due to the simplicity of our network architecture, our method exhibits advantages in inference time and parameters compared to other methods. For detailed information, please refer to the supplementary file.

\begin{table}[!t]
 \setlength{\abovecaptionskip}{0pt} 
  \setlength{\belowcaptionskip}{0pt}
  \caption{{User study on evaluating inbetweening performance of competing methods.}
  }
  \label{tab:user study on method}
  \centering
  \begin{tabular}{@{}p{1.5cm}p{1.5cm}p{1.7cm}p{1.7cm}@{}}
    \toprule
     & Ours & AnimeInbet & EMA-VFI\\
    \midrule
     $N=1$ & 57.50\% & 31.50\% & 11.00\% \\
     $N=5$ & 75.17\% & 16.33\% & 8.50\% \\
     $N=9$ & 69.50\% & 16.50\% & 14.00\% \\
     All & \textbf{70.50}\% & 19.40\% & 10.10\% \\
  \bottomrule
  \end{tabular}
\end{table}

\begin{table}[!t]
 \setlength{\abovecaptionskip}{0pt} 
  \setlength{\belowcaptionskip}{0pt}
  \caption{User study on the consistency of visual perception and evaluation metrics.
  }
  \label{tab:user study on metric}
  \centering
  \begin{tabular}{@{}p{1.5cm}p{1.5cm}p{1.5cm}p{1.5cm}@{}}
    \toprule
    & CD &  WCD &  EMD\\
    \midrule
    $N=1$ & 55.07\% & 60.67\% & 52.25\% \\
    $N=5$ & 57.56\% & 77.41\% & 66.57\% \\
    $N=9$ & 28.66\% & 68.90\% & 59.15\% \\
    All & 51.74\% & \textbf{72.50}\% & 61.73\% \\
  \bottomrule
  \end{tabular}
\end{table}

\subsection{User Study}
To further demonstrate the validity of our proposed metrics and assess the visual performance of our method, we conducted a user study involving 20 participants. Each participant was presented with 50 sets of images randomly selected from the results of EMA-VFI, AnimeInbet, and our method. The ratio of frame gap choices was 20\% for 1, 60\% for 5, and 20\% for 9. Participants were tasked with selecting the visual performance they deemed superior. To factor in temporal consistency in participants' decision-making, we presented these sets in the form of GIFs.

As shown in \cref{tab:user study on method}, most participants preferred the results of our method, especially for large motion scenarios. The user study also demonstrates the validity of our proposed metric WCD compared with CD in \cref{tab:user study on metric}. When the frame gap is small, the differences between the different metrics are not obvious. However, when the frame gap is 9, only approximately 29\% of the participants preferred the result with better CD, contrasting with 69\% and 59\% of participants who preferred results with WCD and EMD, respectively.

\subsection{Ablation Study}
We conduct ablations to verify the effectiveness of certain modules and losses, evaluating the ablation study on a validation set with a frame gap of 5.

\begin{figure}[tb]
  \centering
   \setlength{\abovecaptionskip}{0pt} 
  \setlength{\belowcaptionskip}{0pt}
  \includegraphics[width=\linewidth]{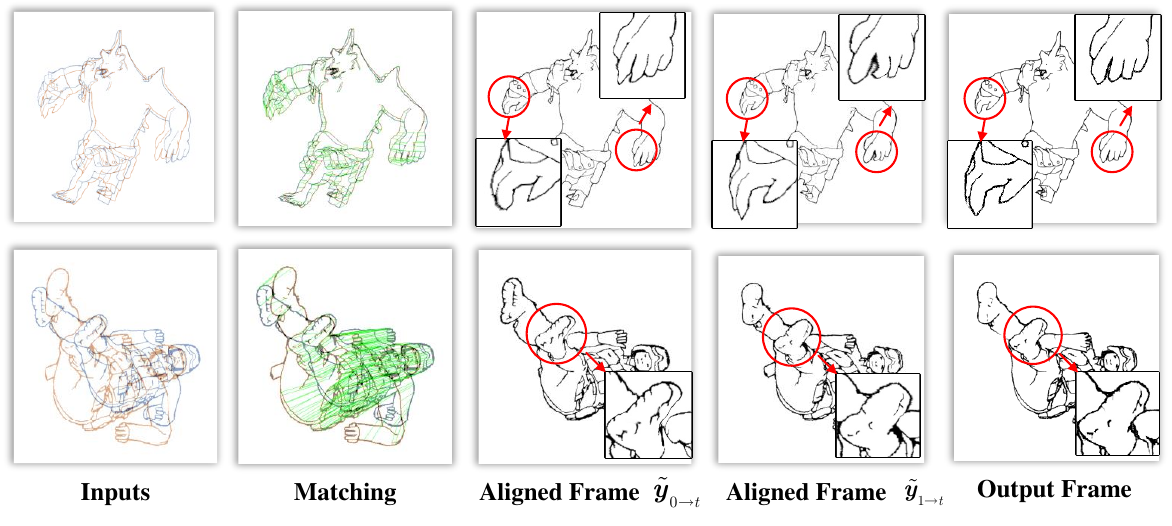}
  \caption{Visual comparison of ablation study on modules, where TPS module can well predict poses in intermediate frames, while FlowNet and UNet bring benefits to detail enhancement.}
  \label{fig:ablation}
\end{figure}

Initially, we verified the effectiveness of the TPS-based coarse motion estimation module that is the core of our approach. As shown in \cref{tab:ablation on module}, `w/o MRM' indicates the utilization of only TPS module for estimating coarse motion, with backward warping applied to generate frames. 
It is observed that employing only coarse motion yields satisfactory results in \cref{fig:ablation}, though the evaluation metrics are not as favorable due to the misalignment with the ground truth. 

In our model, the IFBlock is employed to estimate the motion between two aligned frames. The absence of this flow refinement will result in a decrease in the model's performance as demonstrated in \cref{tab:ablation on module} 'w/o flow refine'. The impact on metrics is not substantial, given that the preceding coarse motion already provides enough information.

More ablation studies on loss functions are available in the supplementary file.

\begin{table}[!t]
 \setlength{\abovecaptionskip}{0pt} 
  \setlength{\belowcaptionskip}{0pt}
  \caption{{Ablation study on modules.}
  }
  \label{tab:ablation on module}
  \centering
  \begin{tabular}{@{}p{2.5cm}p{1cm}p{1cm}p{1cm}@{}}
    \toprule
      & CD &  WCD &  EMD\\
    \midrule
    w/o MRM & 11.800 & 8.850 & 4.233\\
    w/o flow refine & 9.314 & 8.015 & 2.625 \\
    full model & \textbf{9.235}& \textbf{7.994} & \textbf{1.343} \\
  \bottomrule
  \end{tabular}
\end{table}

\section{Conclusion}

In this work, we propose a novel framework for addressing the animation line inbetweening problem based on raster images. By leveraging thin-plate spline-based transformation, more accurate estimates of coarse motion can be obtained by modeling point correspondence between key frames, particularly in scenarios involving significant motion amplitudes. Subsequently, a motion refine module is adopted to enhance motion details and synthesize final frames. 
To mitigate the inherent shortcomings of CD, we introduce an improved metric called WCD that aligns more closely with human visual perception. 
Our quantitative and qualitative evaluations demonstrate the superiority of our method over existing approaches, highlighting its potential impact in relevant fields.

{\small
\bibliographystyle{ieee_fullname}
\bibliography{egbib}
}

\end{document}